**ORIGINAL   ARTICLE**

# Large Scale Local Online Similarity/Distance Learning Framework based on Passive/Aggressive

Baida Hamdan[1], Davood Zabihzadeh*[1], Monsefi Reza[1]

[1]Computer Department, Engineering Faculty, Ferdowsi University of Mashhad (FUM), Mashhad, IRAN

baida.hamdan@mail.um.ac.ir, d.zabihzadeh@mail.um.ac.ir, monsefi@um.ac.ir

* Corresponding Author

## Abstract

Similarity/Distance measures play a key role in many machine learning, pattern recognition, and data mining algorithms, which leads to the emergence of metric learning field. Many metric learning algorithms learn a global distance function from data that satisfy the constraints of the problem. However, in many real-world datasets that the discrimination power of features varies in the different regions of input space, a global metric is often unable to capture the complexity of the task. To address this challenge, local metric learning methods are proposed that learn multiple metrics across the different regions of input space. Some advantages of these methods are high flexibility and the ability to learn a nonlinear mapping but typically achieves at the expense of higher time requirement and overfitting problem. To overcome these challenges, this research presents an online multiple metric learning framework. Each metric in the proposed framework is composed of a global and a local component learned simultaneously. Adding a global component to a local metric efficiently reduce the problem of overfitting. The proposed framework is also scalable with both sample size and the dimension of input data. To the best of our knowledge, this is the first local online similarity/distance learning framework based on PA (Passive/Aggressive). In addition, for scalability with the dimension of input data, DRP (Dual Random Projection) is extended for local online learning in the present work. It enables our methods to be run efficiently on high-dimensional datasets, while maintains their predictive performance. The proposed framework provides a straightforward local extension to any global online similarity/distance learning algorithm based on PA. Experimental results on some challenging datasets from machine vision confirm that the extended methods considerably enhance the performance of related global ones without increasing the time complexity.

**Keywords:** Metric Learning, Local Learning, PA (Passive/Aggressive), Online Learning, Dual Random Projection

## 1. Introduction

Similarity/Distance measures are a key component in many machine learning and data mining algorithms. For example, clustering methods or kNN classifier are built upon a similarity/distance measure. In addition, information retrieval or ranking algorithms need a measure to sort the retrieved items based on degrees of relevancy. However, common measures like Euclidean distance or cosine similarity are not appropriate for many applications. Figure 1 shows an example that Euclidean distance failed to recognize the true relevant image of the target image.

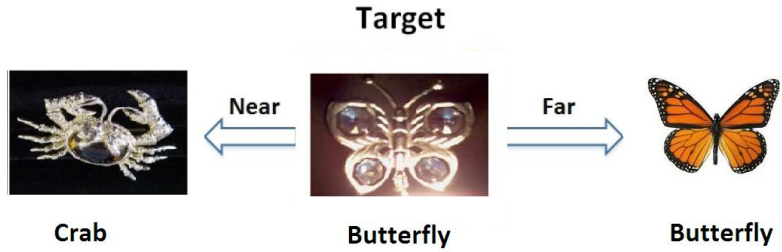

**Figure 1: Problem of Euclidean distance. The left image is from a different class of the target image while under this distance it is closer than the right one to the target image.**

Often a better measure can be learned from data, leading to the emergence of metric learning domain. Metric learning methods aim to learn an optimal distance function from data that closes semantically similar observations while pushes dissimilar ones away from each other. Most methods learns the metric from pair or triplet constraints of the following form:

$S = \{(\boldsymbol{p}_i, \boldsymbol{p}_i^+) \mid \boldsymbol{p}_i \text{ and } \boldsymbol{p}_i^+ \text{ are similar}\}$

$D = \{(\boldsymbol{p}_i, \boldsymbol{p}_i^-) \mid \boldsymbol{p}_i \text{ and } \boldsymbol{p}_i^- \text{ are dissimilar}\}$

$T = \{(\boldsymbol{p}_i, \boldsymbol{p}_i^+, \boldsymbol{p}_i^-) \mid \boldsymbol{p}_i \text{ should be more closer to } \boldsymbol{p}_i^+ \text{ than } \boldsymbol{p}_i^-\}$

Metric learning algorithms can be classified as local and global. In Global approaches a single metric is used to measure the distance between objects. These approaches equivalent to learning a linear or nonlinear projection of data and utilizing the Euclidean distance in this space. However, in many datasets with nonlinear structure and complex class boundary between objects, a single metric is not flexible enough to capture the complexity of the task. To address this challenge, local metric learning methods (Fetaya and Ullman 2015; Shi et al. 2014; Verma et al. 2012; Weinberger and Saul 2009) are proposed that learn multiple metrics across the different regions of input space. These methods often outperform the global ones but this typically achieves at the expense of higher time requirement. These are also more prone to overfitting.

To address these challenges this research presents an online multiple metric learning framework. Each metric in the proposed framework is composed of a global and a local component learned simultaneously. Adding a global component to a local metric efficiently reduces the problem of overfitting. It is worthwhile pointing out that using a common component for multiple hyper-planes was applied successfully in (Zhou et al. 2016). However, unlike this method that is designed for classification task, our work focuses on learning similarity/distance measure.

The proposed framework addresses scalability with respect to sample size by learning the metrics jointly in an online manner. To the best of our knowledge, this is the first local online metric learning method based on PA. In addition, to address scalability with the dimension of input data, DRP (Dual Random Projection) is extended for online learning in the present work. It enables our methods to be run on high-dimensional datasets efficiently, while preserving their predictive performance.

The proposed framework provides a straightforward local extension to any global similarity/distance learning algorithm based on PA. This localization significantly improves the discrimination power of the learned metric without increasing the time complexity of the algorithm. As an example two global similarity/distance learning methods are localized by the proposed framework. Then, the performance of these methods are compared with the local ones for classification task. Extensive evaluations confirm this localization significantly improves the predictive performance of learned metrics.

The major novelty and basic findings of this research can be summarized as follows:

1- Adding a global component to the local metrics to share the discriminating information among them which increases the discrimination power of local metrics and efficiently reduces the problem of overfitting.
2- Local online similarity/distance learning based on PA.
3- Straightforward local extension to any global similarity/distance learning algorithm based on PA which enhances the efficiency of the learned metric without increasing the time complexity.
4- Using DRP in local online similarity/distance learning framework which provides scalability with respect to the dimension of input space while preserving the quality of the learned metric.

The rest of this paper is organized as follows: Section 2 reviews related works. Section 3 presents the proposed framework for online multiple similarity/distance learning based on PA. Experimental results are reported in Section 4. Finally, Section 5 concludes with remarks and recommendations for future work.

## 2. Related Works

Data is being generated in huge quantities every day. Social networks like Facebook and Twitter generate large volumes of data and Google even more. Efficient algorithms are needed to process these huge amounts of data. Traditional batch methods take a lot of time for processing data and are not applicable on large datasets. On the other hand, online algorithms are capable of rapidly process massive datasets with millions of records.

PA is one of the best online learning methods used in various applications. One of the main advantages of this method is that it adaptively adjust the learning rate. PA was first introduced in (Crammer et al. 2006) for classification task. Subsequently, it was used in other applications such as online local categorization (Zhou et al. 2016), recommender systems (Blondel et al., 2014), similarity learning and visual search (Chechik et al., 2010; Wu et al. 2016; Xia et al. 2014; Zhong et al., 2016). In the following the previous works for online similarity/distance learning based on PA are reviewed.

**OASIS[1](Chechik et al. 2010)**
OASIS is an online scalable similarity learning method that efficiently can be run on large datasets with over 2 million images. The training data is assumed to be a triplet of images $(\boldsymbol{p}_i, \boldsymbol{p}_i^+, \boldsymbol{p}_i^-)$ arrives sequentially. It learns a similarity matrix $\boldsymbol{W}$ by minimizing the following optimization problem.

---

[1] <u>O</u>nline <u>A</u>lgorithm for <u>S</u>calable <u>I</u>mage <u>S</u>imilarity

$$W_{t+1} = \arg\min \frac{1}{2}\|W - W_t\|_F^2 + c\xi \qquad (1)$$
$$1 - S_W(p_t, p_t^+) + S_W(p_t, p_t^-) \leq \xi, \ \xi \geq 0$$

where $S_W(p, q)$ is the similarity measure of $p$ and $q$ is defined as:

$$S_W(p, q) = p^T W q \qquad (2)$$

Contrary to metric learning methods, In OASIS, the p.s.d (positive semi-definite) constraint of similarity matrix has been eliminated for speed and scalability. The introduced similarity function has two advantages: 1) Computing similarity of sparse data is very fast. 2) Unlike the Distance learning methods, the comparative data can have different dimensions.

**OMKS [1](Xia et al. 2014)**

Many metric learning methods such as (Davis et al. 2007; Harandi et al. 2017; Weinberger and Saul 2009; Zadeh et al. 2016) assume that the optimal distance function follows the form of general Mahalanobis distances. Also, most of them are not suitable for multi-modal data that has multiple sources of features. To overcome these limitations, in OMKS, the OASIS method was first extended in kernel mode then a multiple kernel learning algorithm is developed based on it.

The similarity function in kernel mode is defined as $S_L(p, q) = \phi(p)^T L \phi(q)$ where $\phi$ is the feature-map induced by a kernel function. To find the optimal linear operator $L$ in feature-space, the following optimization problem based on PA is suggested:

$$L_{t+1} = \arg\min \frac{1}{2}\|L - L_t\|_{HS}^2 + c\xi \qquad (3)$$
$$1 - S_L(p_t, p_t^+) + S_L(p_t, p_t^-) \leq \xi, \ \xi \geq 0$$

Let $k_1, k_2, \ldots, k_m$ are kernel functions, and the vector $\theta = [\theta_1, \theta_2, \ldots, \theta_m]^T$ specifies the coefficients of these functions with conditions ($\theta_i \geq 0$, $\sum_{i=1}^m \theta_i = 1$). The similarity function in multiple kernel mode is defined as follows:

$$f(p, q) = \sum_{i=1}^m \theta_i S_i(p, q) = \sum_{i=1}^m \theta_i \phi_i(p)^T L_i \phi_i(q) \qquad (4)$$

In this case, the optimization problem is formulated as follows:

$$\min_\theta \min_{L_i} \sum_{i=1}^m \theta_i \|L_i\|_{HS}^2 + C \sum_{i=1}^T \ell(p_i, p_i^+, p_i^-) \qquad (5)$$

where $\ell$ is the margin-based hinge loss function.

---

[1] Online Multiple Kernel Similarity Learning

To solve the above problem, the operator $L$ is optimized for each kernel function separately, and then the kernel function coefficients are updated using the Hedge algorithm (Freund and Schapire 1997). The OMKS method has been tested successfully on image data sets.

**OMDML[1] (Wu et al. 2016)**

Both OASIS and OMKS have been eliminated the p.s.d constraint for scalability with respect to the dimension of feature space, while this constraint is very efficient to avoid overfitting and to produce a low-rank solution. In OMDML, the p.s.d constraint has again been added to the optimization problem. OMDML is a multi-metric learning method for multimodal data (data with multiple sources or features). In this method, the ODML algorithm is utilized as the basis for learning each metric. ODML[2] is formulated as follows:

$$M_{t+1} = \arg\min_{M} \frac{1}{2} \|M - M_t\|_F^2 + c\xi \tag{6}$$
$$1 + d_M(p_t, p_t^+) - d_M(p_t, p_t^-) \leq \xi, \ \xi \geq 0, \ M \succcurlyeq 0$$

To enforce the p.s.d constraint, full eigenvalue decomposition has been used at each stage which requires $O(d^3)$ computational complexity. Hence, this method is not scalable in terms of data dimensions and cannot be directly executed on high-dimensional datasets with thousands of features. To solve this problem in (Wu et al., 2016), the dimensions of the data were initially reduced by the PCA. However, dimensionality reduction methods like PCA are unable to take supervised information into account and as a result, this reduction will degrade the accuracy of the learned metric (Qian et al. 2015).

**SLMOML[3] (Zhong et al. 2016)**

SLMOML uses $logdet$ regularization term instead of Frobenius norm to automatically ensure the p.s.d property of the similarity matrix at each iteration. Therefore, there is no need to eigenvalue decomposition in this method. Although this method is more scalable with respect to the dimensionality of data than previous approaches, SLMODML still needs to maintain $O(d^2)$ parameters and also suffers from low convergence rate.

As can be seen, existing methods for learning similarity/distance based on PA are global, and none of them takes the local distribution of data into consideration which leads to learn a more effective similarity/distance measure. In addition, scalability with respect to the dimension of data is a major challenge in this domain. The proposed solutions to overcome these challenges are presented in the next section.

## 3. Online Multi-Metric Similarity/Distance Learning Framework

In this section, the proposed framework for online multi-metric similarity/distance learning is discussed. Each metric in the proposed framework is composed of a global and a local component learned jointly. Adding a common component to a local metric efficiently reduces the problem of overfitting. This framework provides a straightforward local extension to any global similarity/distance learning algorithm based on PA. For example ODML (Wu et al. 2016) and OASIS (Chechik et al. 2010) have been extended by applying the proposed framework and two new local online similarity/distance learning algorithms are produced. These methods are named LOSL[1] and LODML[2] respectively.

---

[1] Online Multi-Modal Distance Metric Learning

[2] Online Distance Metric Learning

[3] Scalable Large Margin Online Metric Learning

These methods are scalable due to online data processing with respect to the size of the dataset. To provide scalability with the dimension of input data, DRP approach is first extended for online similarity/distance learning and then this approach is applied to both LODML and LOSL methods. The resulting algorithms are named L2OSL[3] and L2ODML[4] which are scalable with both sample size and the dimension of data.

## 3.1 LODML

This method is the local extension of ODML. LODML defines a metric $M_j$ for each class $j$ in the dataset as follows:

$$M_j = \lambda U + \mu L_j \quad j = 1,2,\ldots,c \quad \lambda \in [0,1], \ \mu = (1-\lambda) \tag{7}$$

In the above relation, $c$ is the number of classes in dataset, $U$ is the common metric between all classes, and $L_j$ is the local metric associated to class $j$. Hence, the metric of each class benefits of both the global and local discriminative information. The coefficient $\lambda$ controls the tradeoff between them. Adding a common component to a local metric efficiently reduces the problem of overfitting. The extensive experimental results in the next section confirms the truth of this statement.

The distance function of metric $M_j$ is as follows:

$$d_{M_j}(p,q) = (p-q)^T(\lambda U + \mu L_j)(p-q) \tag{8}$$

Let triple set $P = \{(p_i, p_i^+, p_i^-) | p_i \text{ should be more similar to } p_i^+ \text{ than to } p_i^-\}$ represent the training set arrived sequentially to the algorithm. At time step $k$, a triplet $(p_k, p_k^+, p_k^-)$ is entered to the system. Let $p_k, p_k^+$ belongs to class $i$, and $j$ is the label of $p_k^-$. The following optimization problem is proposed to learn the metrics.

$$U^{(k+1)}, L_i^{(k+1)}, L_j^{(k+1)} = \arg\min_{U,L_i,L_j} \frac{1}{2}\|U - U^{(k)}\|_F^2 + \frac{1}{2}\|L_i - L_i^{(k)}\|_F^2 + \frac{1}{2}\|L_j - L_j^{(k)}\|_F^2 + C\xi$$

$$subject\ to \tag{9}$$

$$1 + d_{M_i}(p_k, p_k^+) - d_{M_j}(p_k, p_k^-) \leq \xi, \ \xi \geq 0, \ U, L_i, L_j \succcurlyeq 0$$

---

[1] Local Online Similarity Learning

[2] Local Online Distance Metric Learning

[3] Large Scale Local Similarity Learning

[4] Large Scale Online Local Distance Metric Learning

Hence, at each iteration $k$, the global metric $U$ and the local metrics $L_i$ and $L_j$ are updated to optimize the trade-off between minimizing the margin-based hinge loss on the current triplet $(p_k, p_k^+, p_k^-)$ and also remaining close to the previous values to avoid overfitting. The balance of this trade-off is controlled by the parameter $C$.

**Theorem1**: The relation (9) admits a closed-form solution which can be expressed as:

$$U^{(k+1)} = U^{(k)} - \tau\lambda G_d$$
$$L_i^{(k+1)} = L_i^{(k)} - \tau\mu G^+ \qquad (10)$$
$$L_j^{(k+1)} = L_j^{(k)} + \tau\mu G^-$$

where

$$G^+ = (p_k - p_k^+)(p_k - p_k^+)^T, \quad G^- = (p_k - p_k^-)(p_k - p_k^-)^T$$
$$G_d = G^+ - G^- \qquad (11)$$

and also the learning rate parameter $\tau$ is as follows:

$$\tau = \min\left(C, \frac{l^{(k)}(p_k, p_k^+, p_k^-)}{\|\lambda G_d\|_F^2 + \|\mu G^+\|_F^2 + \|\mu G^-\|_F^2}\right) \qquad (12)$$

where $l^{(k)}(p_k, p_k^+, p_k^-) = \max\left(0, 1 + \lambda \operatorname{tr}(U^{(k)} G_d) + \mu \operatorname{tr}(L_i^{(k)} G^+) - \mu \operatorname{tr}(L_j^{(k)} G^-)\right)$ shows the value of loss function in the current model (before the update).

***Proof***: The proof of this theorem is provided in Appendix A. ∎

According to the above results, the pseudo-code of ODML is summarized in Algorithm 1.

### 3.2 LOSL

As stated, the proposed framework easily can be applied to any similarity/distance learning based on PA. In this subsection, the local extension of OASIS algorithm is provided. The resulting local online similarity learning algorithm is named LOSL.

LOSL defines a similarity matrix $M_j$ for each class $j$ in the dataset as follows:

$$M_j = \lambda U + \mu L_j \quad j = 1, 2, \ldots, c \quad \lambda \in [0,1], \ \mu = (1 - \lambda) \qquad (13)$$

In the above relation, $U$ is the similarity matrix shared between all classes and $L_j$ is the local similarity matrix associated to class $j$. Hence, the similarity matrix $M_j$ benefits of both the global and local discriminative information. The coefficient $\lambda$ balances the tradeoff between them.

The similarity function of $M_j$ is as follows:

$$S_{M_j}(p, q) = p^T(\lambda U + L_j)q = \lambda p^T U q + \mu\, p^T L_j q \quad j = 1, 2, \ldots, c \qquad (14)$$

**Algorithm1.** LODML- Local Online Distance Metric Learning

---
Input: C: aggressiveness parameter

    1. Initialize the global matrix $U$ and the local matrices $L_i$ $i = 1,2, \ldots, c$ with Identity matrix.

    2. for $k = 1,2, \ldots, T$

        2.1. receive a triplet $(p_k, p_k^+, p_k^-)$

        2.2. $G^+ = (p_k - p_k^+)(p_k - p_k^+)^T, \quad G^- = (p_k - p_k^-)(p_k - p_k^-)^T$

            $G_d = G^+ - G^-$

        2.3. $\tau = \min\left(C, \frac{l^{(k)}(p_k, p_k^+, p_k^-)}{\|\lambda G_d\|_F^2 + \|\mu G^+\|_F^2 + \|\mu G^-\|_F^2}\right)$

        2.4. $U = U^{(k)} - \tau \lambda G_d$

        2.5. $L_i = L_i^{(k)} - \tau \mu G^+ \quad i$ is a label of $p_k$

        2.6. $L_j = L_j^{(k)} + \tau \mu G^- \quad j$ is a label of $p_k^-$

    end;

    3. $U = \text{psd}(U)$.

    4. for $i = 1,2, \ldots, c \quad c$: the number of classes in dataset

        4.1. $L_i = \text{psd}(L_i)$.

---

As before, assume the training set $P = \{(p_i, p_i^+, p_i^-) | p_i \text{ should be more similar to } p_i^+ \text{ than to } p_i^-\}$ enters sequentially to the model. Let $p_k, p_k^+$ belongs to class $i$ and $j$ is the label of $p_k^-$. The similarity function are optimized by minimizing the following optimization problem.

$$U^{(k+1)}, L_i^{(k+1)}, L_j^{(k+1)} = \arg\min_{U, L_i, L_j} \frac{1}{2}\|U - U^{(k)}\|_F^2 + \frac{1}{2}\|L_i - L_i^{(k)}\|_F^2 + \frac{1}{2}\|L_j - L_j^{(k)}\|_F^2 + C\xi \quad (15)$$

*subject to*

$$1 - S_{M_i}(p_k, p_k^+) + S_{M_j}(p_k, p_k^-) \leq \xi, \ \xi \geq 0, \ U, L_i, L_j \succcurlyeq 0$$

To solve the above problem, note that

$$S_i(p_k, p_k^+) = p_k^T(\lambda U + L_i)p_k^+ = \lambda \, \text{tr}(U p_k^+ p_k^T) + \mu \, \text{tr}(L_i p_k^+ p_k^T)$$

and

$$S_j(p_k, p_k^-) = p_k^T(\lambda U + L_j)p_k^- = \lambda \, \text{tr}(U p_k^- p_k^T) + \mu \, \text{tr}(L_j p_k^- p_k^T)$$

Let

$$F^+ = p_k^+ p_k^T, \quad F^- = p_k^- p_k^T$$
$$F_d = F^+ - F^- \quad (16)$$

Then, the above expressions can be written as follows:

$$S_i(\boldsymbol{p}_k, \boldsymbol{p}_k^+) = \lambda \operatorname{tr}(\boldsymbol{UF}^+) + \mu \operatorname{tr}(\boldsymbol{L}_i \boldsymbol{F}^+)$$
$$S_j(\boldsymbol{p}_k, \boldsymbol{p}_k^-) = \lambda \operatorname{tr}(\boldsymbol{UF}^-) + \mu \operatorname{tr}(\boldsymbol{L}_j \boldsymbol{F}^-)$$

and we have

$$\begin{aligned} 1 - S_{M_i}(\boldsymbol{p}_k, \boldsymbol{p}_k^+) + S_{M_j}(\boldsymbol{p}_k, \boldsymbol{p}_k^-) &= 1 - \lambda \operatorname{tr}(\boldsymbol{UF}^+) - \mu \operatorname{tr}(\boldsymbol{L}_i \boldsymbol{F}^+) + \lambda \operatorname{tr}(\boldsymbol{UF}^-) + \mu \operatorname{tr}(\boldsymbol{L}_j \boldsymbol{F}^-) \\ &= 1 - \lambda \operatorname{tr}(\boldsymbol{UF}_d) - \mu \operatorname{tr}(\boldsymbol{L}_i \boldsymbol{F}^+) + \mu \operatorname{tr}(\boldsymbol{L}_j \boldsymbol{F}^-) \end{aligned} \tag{17}$$

So according to the solution provided for (9) in Appendix A, the solution of (15) is as follows:

$$\begin{aligned} \boldsymbol{U}^{(k+1)} &= \boldsymbol{U}^{(k)} + \tau \lambda \boldsymbol{F}_d^T \\ \boldsymbol{L}_i^{(k+1)} &= \boldsymbol{L}_i^{(k)} + \tau \mu \boldsymbol{F}^{+T} \\ \boldsymbol{L}_j^{(k+1)} &= \boldsymbol{L}_j^{(k)} - \tau \mu \boldsymbol{F}^{-T} \end{aligned} \tag{18}$$

and also the step size parameter $\tau$ is:

$$\tau = \min\left(C, \frac{l^{(k)}(\boldsymbol{p}_k, \boldsymbol{p}_k^+, \boldsymbol{p}_k^-)}{\|\lambda \boldsymbol{G}_d\|_F^2 + \|\mu \boldsymbol{G}^+\|_F^2 + \|\mu \boldsymbol{G}^-\|_F^2}\right) \tag{19}$$

where $l^{(k)}(\boldsymbol{p}_k, \boldsymbol{p}_k^+, \boldsymbol{p}_k^-) = \max\left(0, 1 - \lambda \operatorname{tr}(\boldsymbol{U}^{(k)} \boldsymbol{F}_d) - \mu \operatorname{tr}\left(\boldsymbol{L}_i^{(k)} \boldsymbol{F}^+\right) + \mu \operatorname{tr}\left(\boldsymbol{L}_j^{(k)} \boldsymbol{F}^-\right)\right)$ shows the value of loss function in the current model (before the update).

Note that the time complexity of LODML and LOSL is the same as corresponding global methods. The *dominant operation* in the local methods is the computation of $\boldsymbol{G}_d$ which also exists in the global ones.

### 3.3 Scalability with the dimension

The LODML and LOSL methods due to online data processing efficiently can be run on large datasets. However, these methods are not scalable with respect to the dimension of data. To overcome this problem, DRP approach is adopted in the proposed framework. In this subsection, we show how DRP can be applied to any online learning algorithm in the proposed framework. As an example, LODML and LOSL methods are extended by applying this approach. The resulting algorithms are named L2OSL and L2ODML which are scalable with both sample size and the dimension of data.

**L2ODML**

In DRP, First a random matrix $\boldsymbol{R} \in \mathbb{R}^{d \times m}$ is generated where $m \ll d$ and $R_{i,j} = \mathcal{N}(0, 1/m)$, then the data points are projected into low-dimensional subspace using this matrix, i.e., $\hat{\boldsymbol{p}}_k = \boldsymbol{R}^T \boldsymbol{p}_k$ $k = 1, 2, \ldots, T$. As a result, the matrices $\boldsymbol{G}^+$, $\boldsymbol{G}^-$, and $\boldsymbol{G}_d$ after projection become

$$\hat{\boldsymbol{G}}_d = \boldsymbol{R}^T \boldsymbol{G}_d \boldsymbol{R}, \quad \hat{\boldsymbol{G}}_d \in \mathbb{R}^{m \times m}$$
$$\hat{\boldsymbol{G}}^+ = \boldsymbol{R}^T \boldsymbol{G}^+ \boldsymbol{R}, \quad \boldsymbol{G}^+ \in \mathbb{R}^{m \times m}$$
$$\hat{\boldsymbol{G}}^- = \boldsymbol{R}^T \boldsymbol{G}^- \boldsymbol{R}, \quad \boldsymbol{G}^- \in \mathbb{R}^{m \times m}$$

Also, the optimization problem (9) in the low-dimensional subspace is as follows:

$$\widehat{U}^{(k+1)}, \widehat{L}_i^{(k+1)}, \widehat{L}_j^{(k+1)} = \arg\min_{\widehat{U},\widehat{L}_i,\widehat{L}_j} \frac{1}{2}\|\widehat{U} - \widehat{U}^{(k)}\|_F^2 + \frac{1}{2}\|\widehat{L}_i - \widehat{L}_i^{(k)}\|_F^2 + \frac{1}{2}\|\widehat{L}_j - \widehat{L}_j^{(k)}\|_F^2 + C\xi \qquad (20)$$

subject to

$$1 + \lambda \operatorname{tr}(\widehat{U}\widehat{G}_d) + \mu \operatorname{tr}(\widehat{L}_i\widehat{G}^+) - \mu \operatorname{tr}(\widehat{L}_j\widehat{G}^-) \le \xi, \ \xi \ge 0, \ \widehat{U},\widehat{L}_i,\widehat{L}_j \succcurlyeq 0$$

The main limitation of the above optimization problem is that the range space of $\widehat{U}, \widehat{L}_i$, and $\widehat{L}_j$ is limited to the subspace spanned by the columns of random matrix $R$. So instead of solving the primal problem, the dual problem in the projected subspace is solved by DRP approach.

$$\begin{aligned}
\operatorname*{maximize}_{\hat{\tau}} D(\hat{\tau}) &= -\frac{\tau^2}{2}\left(\|\lambda\widehat{G}_d\|_F^2 + \|\mu\widehat{G}^+\|_F^2 + \|\mu\widehat{G}^-\|_F^2\right) \\
&\quad + \tau\left(1 + \lambda\operatorname{tr}(\widehat{U}^{(k)}\widehat{G}_d) + \mu\operatorname{tr}(\widehat{L}_i^{(k)}\widehat{G}^+) - \mu\operatorname{tr}(\widehat{L}_j^{(k)}\widehat{G}^-)\right)
\end{aligned} \qquad (21)$$

subject to $\quad 0 \le \hat{\tau} \le C$

The solution of this problem is as follows:

$$\hat{\tau} = \min\left(C, \frac{\hat{l}^{(k)}(\widehat{p}_k, \widehat{p}_k^+, \widehat{p}_k^-)}{\|\lambda\widehat{G}_d\|_F^2 + \|\mu\widehat{G}^+\|_F^2 + \|\mu\widehat{G}^-\|_F^2}\right) \qquad (22)$$

where $\hat{l}^{(k)}(\widehat{p}_k, \widehat{p}_k^+, \widehat{p}_k^-) = \max\left(0, 1 + \lambda\operatorname{tr}(U^{(k)}\widehat{G}_d) + \mu\operatorname{tr}(L_i^{(k)}\widehat{G}^+) - \mu\operatorname{tr}(L_j^{(k)}\widehat{G}^-)\right)$ is the value of loss function in the current model (before the update).

After obtaining the optimal value of $\hat{\tau}$, the metrics will be updated in the original space as follows:

$$\begin{aligned}
U &= U^{(k)} - \hat{\tau}\lambda G_d \\
L_i &= L_i^{(k)} - \hat{\tau}\mu G^+ \\
L_j &= L_j^{(k)} + \hat{\tau}\mu G^-
\end{aligned} \qquad (23)$$

Note that in this method, the learned matrices are not limited to the subspace spanned by the columns of a random matrix, a key factor for the success of the DRP approach. In addition, computing the learning rate from the relation (22) just requires $O(m^2)$ operations which is considerably more efficient than the relation (12) that needs $O(d^2)$ operations.

The matrices $U$ and $L_i \ i = 1,2,\ldots,c$ have $d^2$ parameters. Therefore, it is almost impossible to store these matrices in a very high-dimensional environment. To overcome this limitation, DRP stores the low-rank representations of these matrices. More specifically, let $X = [x_1, x_2, \ldots, x_n]$ be the set of i.i.d input data from which the triplet set $P$ is constructed. Note that the number of triplets is usually larger than the number of elements in $X$ (i.e., $n \ll N$). Also, let $i_1$, $i_2$, and $i_3$ denote the indices of $p_k$, $p_k^+$ and $p_k^-$ in $X$ respectively. Instead of saving matrix $U$, a sparse matrix $S \in \mathbb{R}^{n \times n}$ is considered and, at each step after determining $\hat{t}$, it is updated as follows:

$$S(i_1, i_2) = S(i_1, i_2) - \hat{t}, \quad S(i_2, i_1) = S(i_2, i_1) - \hat{t}, \quad S(i_2, i_2) = S(i_2, i_2) + \hat{t}, \quad (24)$$
$$S(i_1, i_3) = S(i_1, i_3) + \hat{t}, \quad S(i_3, i_1) = S(i_3, i_1) + \hat{t}, \quad S(i_3, i_3) = S(i_3, i_3) - \hat{t},$$

Note that

$$\begin{aligned} G_d &= (p_k - p_k^+)(p_k - p_k^+)^T - (p_k - p_k^-)(p_k - p_k^-)^T \\ &= (x_{i_1} - x_{i_2})(x_{i_1} - x_{i_2})^T - (x_{i_1} - x_{i_3})(x_{i_1} - x_{i_3})^T \\ &= -x_{i_1} x_{i_2}^T - x_{i_2} x_{i_1}^T + x_{i_2} x_{i_2}^T + x_{i_1} x_{i_3}^T + x_{i_3} x_{i_1}^T - x_{i_3} x_{i_3}^T \end{aligned}$$

Therefore, it is easy to verify that $U$ can be expressed as

$$U = I + XSX^T \qquad (25)$$

Similarly, instead of storing the matrices $L_i$ and $L_j$, we consider sparse matrices $S_i$ and $S_j$. Then, at each step these matrices are updated as follows:

$$\begin{aligned} S_i(i_1, i_1) &= S_i(i_1, i_1) + \hat{t}, \quad S_i(i_1, i_2) = S_i(i_1, i_2) - \hat{t}, \quad S_i(i_2, i_1) = S_i(i_2, i_1) - \hat{t}, \\ S_i(i_2, i_2) &= S_i(i_2, i_2) + \hat{t}, \\ S_j(i_1, i_1) &= S_j(i_1, i_1) - \hat{t}, \quad S_j(i_1, i_3) = S_j(i_1, i_3) + \hat{t}, \quad S_j(i_3, i_1) = S_j(i_3, i_1) + \hat{t}, \\ S_j(i_3, i_3) &= S_j(i_3, i_3) - \hat{t}, \end{aligned} \qquad (26)$$

In the proposed framework, projecting the metrics to the cone of p.s.d. matrices is done by exploiting the randomized theory (Halko et al. 2011) which is presented in Algorithm 1 of (Qian et al. 2015). This method takes the desired matrix like $U$ as an input and returns a matrix $W$ which is the low-rank representation of $U$ ($U \approx WW^T$). The time complexity of this method is $O(ndq)$ where the parameter $q$ should be adjusted in the range ($r < q \ll d$).

The steps of L2ODML is summarized in Algorithm 2.

**Algorithm 2**. L2ODML- Large Scale Local Online Distance Metric Learning

---

Input: C: aggressiveness parameter

    1. Initialize global Matrix $U$ and local metrices $L_i\ i = 1,2,\dots,c$ with Identity matrix

    2. Generate a random matrix $R \in \mathbb{R}^{d \times m}$ where $R_{i,j} = \mathcal{N}(0, 1/m),\ m \ll d$

    3. for $k = 1,2,\dots,T$

        3.1. receive a triplet $(p_k, p_k^+, p_k^-)$

        3.2. project $(p_k, p_k^+, p_k^-)$ to the low-dimensional subspace spanned by columns of $R$,
$$(\hat{p}_k, \hat{p}_k^+, \hat{p}_k^-) = (R^T p_k, R^T p_k^+, R^T p_k^-)$$

        3.3. $\hat{G}^+ = (\hat{p}_k - \hat{p}_k^+)(\hat{p}_k - \hat{p}_k^+)^T,\quad \hat{G}^+ = (\hat{p}_k - \hat{p}_k^-)(\hat{p}_k - \hat{p}_k^-)^T$
$$\hat{G}_d = G^+ - G^-$$

        3.4. $\hat{\tau} = \min\left(C, \dfrac{\hat{l}^{(k)}(\hat{p}_k, \hat{p}_k^+, \hat{p}_k^-)}{\|\lambda \hat{G}_d\|_F^2 + \|\mu \hat{G}^+\|_F^2 + \|\mu \hat{G}^-\|_F^2}\right)$

        3.5. Update matrix $S$ using (24)

        3.6. Update matrix $S_i$ using (26)

        3.7. Update matrix $S_j$ using (26)

    end;

    4. $W =$ Recover the low-rank approximation of $I + XSX^T$ using Algorithm 1 of (Qian et al. 2015)

    5. for $i = 1,2,\dots,c\quad c$: *the number of classes in dataset*

        5.1. $W_i =$ Recover the low-rank approximation of $I + XS_iX^T$ using Algorithm 1 of (Qian et al. 2015)

---

**L2OSL**

The L2OSL algorithm can be similarly implemented. The difference is that now the update rules for the matrices $S$, $S_i$, and $S_j$ are:

$$\begin{aligned}
&S(i_1, i_2) = S(i_1, i_2) + \hat{\tau}, \quad S(i_1, i_3) = S(i_1, i_3) - \hat{\tau}, \\
&S_i(i_1, i_2) = S_i(i_1, i_2) + \hat{\tau} \\
&S_j(i_1, i_3) = S_j(i_1, i_3) - \hat{\tau}
\end{aligned} \qquad (27)$$

## 4. Experimental Results

In this Section, the experiments conducted to confirm the effectiveness of the proposed methods are explained.

### 4.1 Data Description

Most of the evaluated datasets are challenging ones in computer vision. The statistics and brief description of these datasets are reported in Table 1. In this table, #$dim$ shows the original dimension of the dataset and $d$ represents the number of attributes after dimensionality reduction using the PCA or feature extraction with the CNN-Reset152 network[1]. Also, Figure 2 shows random images from the Caltech101, Caltech256, CIFAR10, and Oxford Cats & Dogs datasets.

Table 1-Statistics and brief description of evaluated datasets in our experiments

| Data Set | #classes | #samples | #dim | Feature extraction | d | Description |
|---|---|---|---|---|---|---|
| Letters (Lichman 2013) | 26 | 20,000 | 16 | None | 16 | This dataset include 20,000 instances of 26 English capital letters. Images of letters were generated from 20 different fonts, then 16 primitive numerical attributes were extracted from these images. |
| YaleFaceB (Lee et al. 2005) | 38 | 2,414 | 1,024 | PCA | 200 | This is a standard face recognition dataset which contains $n = 2,414$ images of 38 classes. For each person, 64 images were taken under extreme illumination conditions. Some people have fewer images. |
| Caltech101 (Li et al. 2004) | 101 | 9,146 | variable | CNN Resnet152 | 1000 | Caltech101 considered as a standard image classification dataset. The images belong to 101 classes. Each category contains about 40 to 800 images and the size of each image is roughly 300×200 pixels. |
| Caltech256 (Fei-Fei et al. 2007): | 256 | 30,607 | variable | CNN Resnet152 | 1000 | Caltech256 is a challenging machine vision dataset that contains 30,607 images. This dataset is made by downloading examples from Google Images. |
| CIFAR10 | 10 | 60,000 | 1,024 | CNN Resnet152 | 1000 | CIFAR-10 contains 60,000 32×32 color images. This collection contains 10 classes, and each class includes 6,000 images. The pictures are divided into five training batches and one test batch. There are 10,000 images in the test batch and the training batches contain the remaining instances. |
| Oxford Cats&Dogs (Parkhi et al. 2012) | 37 | 7,349 | variable | CNN Resnet152 | 1000 | In this dataset, the target is to detect animal species from the image. There are 7,349 images of 37 different breeds of cats and dogs. The problem of recognizing races from the image in this dataset is very challenging, especially in the case of cats, because there is little difference between their different breeds in appearance. |

## 4.2 Experimental Setup

Since the normalization of the data in the Letters dataset improves the results for all the competing methods, this dataset is normalized by setting the mean of each feature equal to zero and its variance equal to one. In the YaleFaceB dataset, to reduce the noise effect, the dimensionality of images were reduced to 200 using the PCA. In images datasets, Caltech101, Caltech256, CIFAR10, and Oxford Cats&Dogs, the feature extraction is done using the CNN-Reset152 network. For this purpose, the images were initially rescaled to 224×224 size and subtracted from the average image of the network. Then, a total of 1000 features per image were extracted from the layer just before the output layer.

In the Caltech101 dataset, 10 and 20 classes out of the 101 categories are randomly selected to form two sub-sets of data named Caltech101_10, Caltech101_20. Similarly, for the Caltech256 dataset, the same procedure is done and the Caltech256_10 and Caltech256_20 subcategories are generated.

---

[1]downloaded from http://www.vlfeat.org/matconvnet/pretrained/

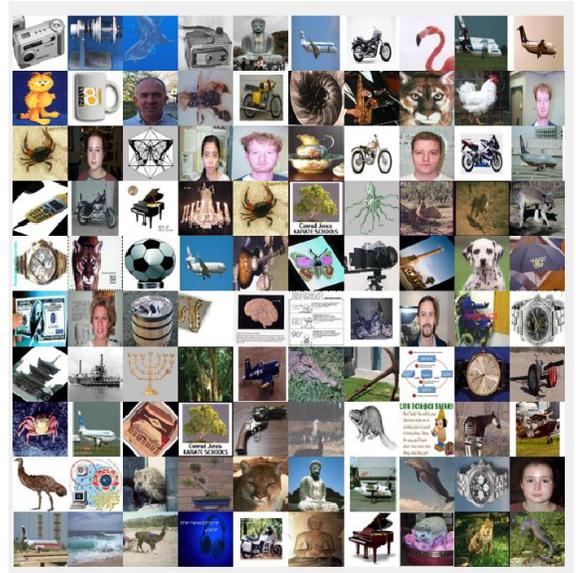
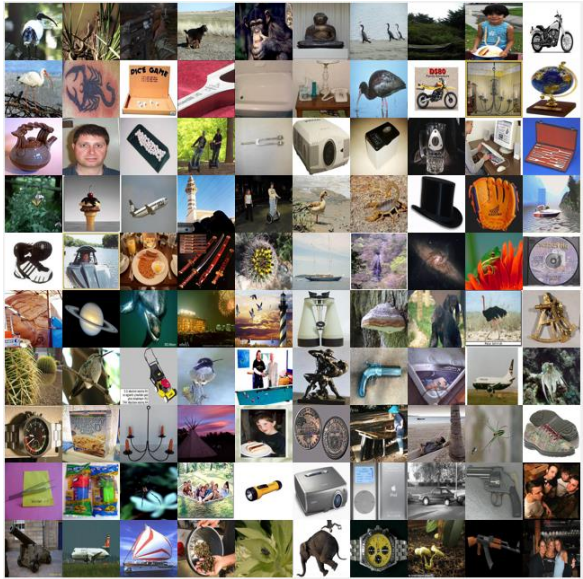

(b)

(a)

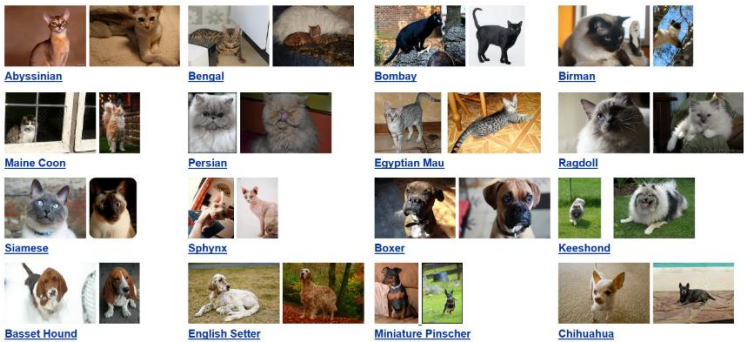
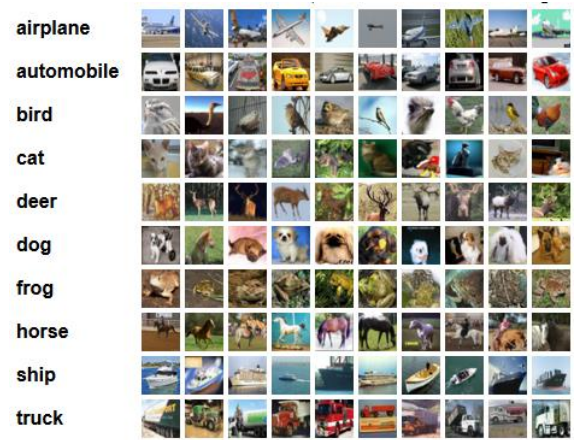

(d)

(c)

**Figure 2: Random images from some evaluated datasets. (a) Caltech 101, (b) Caltech256, (c) CIFAR10, and (d) Oxford Cats&Dogs**

Training and test sets are produced by randomly dividing instances of the datasets at each run, except for CIFAR10 which has a predefined training/test sets. Parameter $f$ in Table 2 indicates the percentage of total examples in each data set utilized for training.

To generate the training triple set in the experiments, each data item $x$ is considered similar to $k$-nearest ($k = 3$) data elements from the same class called *target neighbors* and considered dissimilar to any of its *imposters*. For further explanations of the terms *target neighbors* and *imposters* refer to (Weinberger and Saul 2009). Subsequently, the training triplet set is generated from the Cartesian product of similar and dissimilar pairs.

For a fair comparison, the same triplet set is provided for each method. The kNN classifier $k = 3$ has been adopted to evaluate the efficiency of the learned similarity/distance measures. Figure 3 show the system flow of the similarity/distance learning scheme which consists of two phases namely the learning phase and test phase.

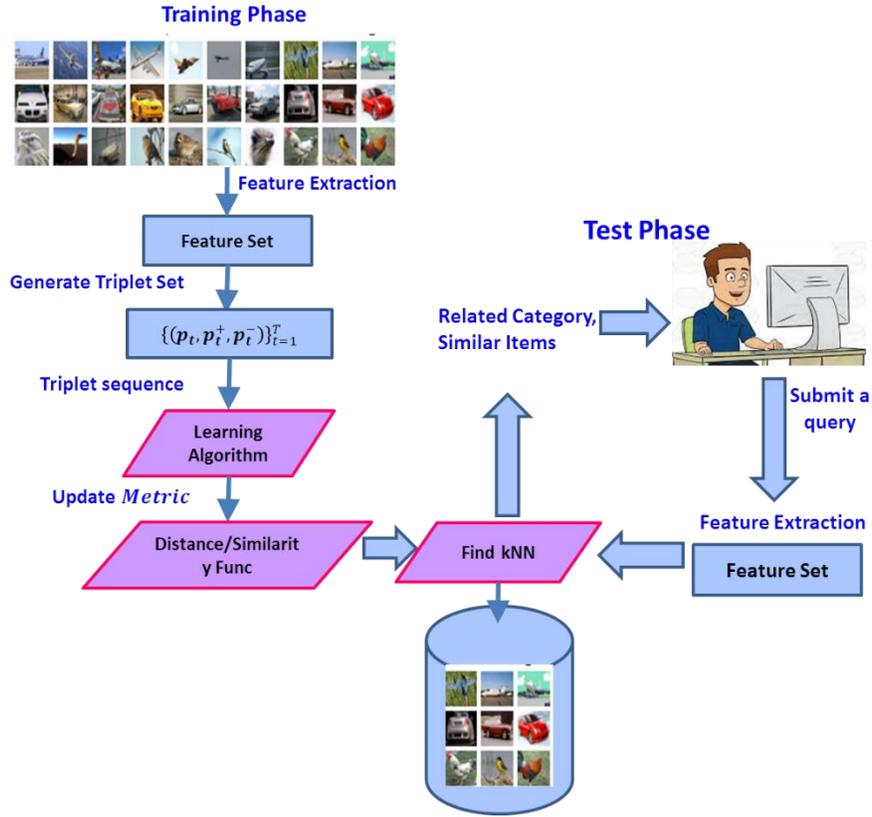

**Figure 3: The system flow of the similarity/distance learning scheme**

In our experiments, the proposed local methods (LODML and LOSL) are compared with the corresponding global ones (ODML and OASIS). The hyperparameters of the competing methods ($C, \lambda$) are adjusted by $k$-fold cross validation with $k = 5$. Also, the efficiency of L2ODM is compared with LODML in terms of both classification rate and execution time.

### 4.3 Results and Analysis

The classification rate of the kNN classifier using the learned similarity/distance measure of the competing methods is presented in Table 2. Experimental results are obtained by averaging over 10 runs on the datasets.

As the results in Table 2 show, localizing of similarity/distance learning algorithms by applying the proposed framework considerably increases the predictive performance of the classifier, and the local LODML and LOSL methods have a significantly higher performance than the corresponding global ones. It can be explained by the fact that each metric in the proposed framework benefits of both the global and local discriminative information. The local component of each metric learns specifically the discriminative features of the associated class which enhances the discrimination power of the learned metric. On the other hand, adding the global component to each local metric has prevented the overfitting.

**Table 2- The classification rate of the kNN classifier using the learned measure of the competing methods**

| Data Set | f % | LODML | ODML | LOSL | OASIS |
|---|---|---|---|---|---|
| Letters | 30 | **94.46+-0.23** | 92.79+-0.53 | 49.86+-1.29 | 49.12+-0.61 |
| | 70 | **96.59+-0.23** | 95.08+-0.41 | 49.94+-1.15 | 48.16+-0.17 |
| YaleFaceB | 30 | 88.33+-0.85 | 92.79+-0.53 | **88.84+-1.05** | 88.17+-0.93 |
| | 70 | **94.46+-0.73** | 95.08+-0.41 | 94.13+-0.72 | 93.44+-1.13 |
| Caltech101_10 | 30 | 98.03+-0.31 | 97.21+-0.55 | **98.28+-0.28** | 98.28+-0.28 |
| | 70 | **98.72+-0.81** | 98.27+-0.36 | 98.66+-0.61 | 98.59+-0.80 |
| Caltech101_20 | 30 | **97.46+-0.38** | 97.02+-0.70 | 97.16+-0.10 | 97.02+-0.40 |
| | 70 | **98.42+-0.65** | 97.85+-0.44 | 98.03+-0.12 | 97.76+-0.15 |
| Caltech256_10 | 30 | 97.67+-0.37 | 97.53+-0.45 | **98.28+-0.39** | 98.02+-0.48 |
| | 70 | 98.58+-0.29 | 98.35+-0.30 | **98.66+-0.26** | 98.23+-0.31 |
| Caltech256_20 | 30 | **96.81+-0.34** | 96.43+-0.24 | 96.08+-0.37 | 96.05+-0.53 |
| | 70 | **97.53+-0.38** | 96.82+-0.47 | 96.40+-0.78 | 95.78+-0.64 |
| CIFAR10 | | **90.57+-0.00** | 90.33+-0.00 | 81.40+-0.00 | 81.40+-0.00 |
| Oxford Cats&Dogs | 30 | **87.21+-0.83** | 82.54+-0.15 | 71.88+-2.21 | 63.06+-2.38 |
| | 70 | **88.96+-0.46** | 86.19+-0.44 | 70.31+-2.51 | 64.10+-1.79 |

In Figure 4, the convergence rate of the competing methods is compared on the evaluated datasets. As can be seen, the convergence rate of the LODML and LOSL is higher than the corresponding global ones. Note that, higher convergence rate in the local methods is achieved while the computational complexity of these methods is the same as the global ones.

In another experiment, the influence of the parameter $\lambda$ on the classification accuracy of the local methods is examined in the Letters and YaleB32 datasets. The parameter $\lambda$ controls that weight of the global and local components for each metric. As the value of $\lambda$ increases, the weight of the local component in each metric decreases and the proposed local methods are reduced to the corresponding global ones when the value of parameter $\lambda$ is equal to one. As shown in Figure 5, the predictive performance of LODML is maximized when both the local and global components are jointly learned and utilized to measure the similarity/distance between data items.

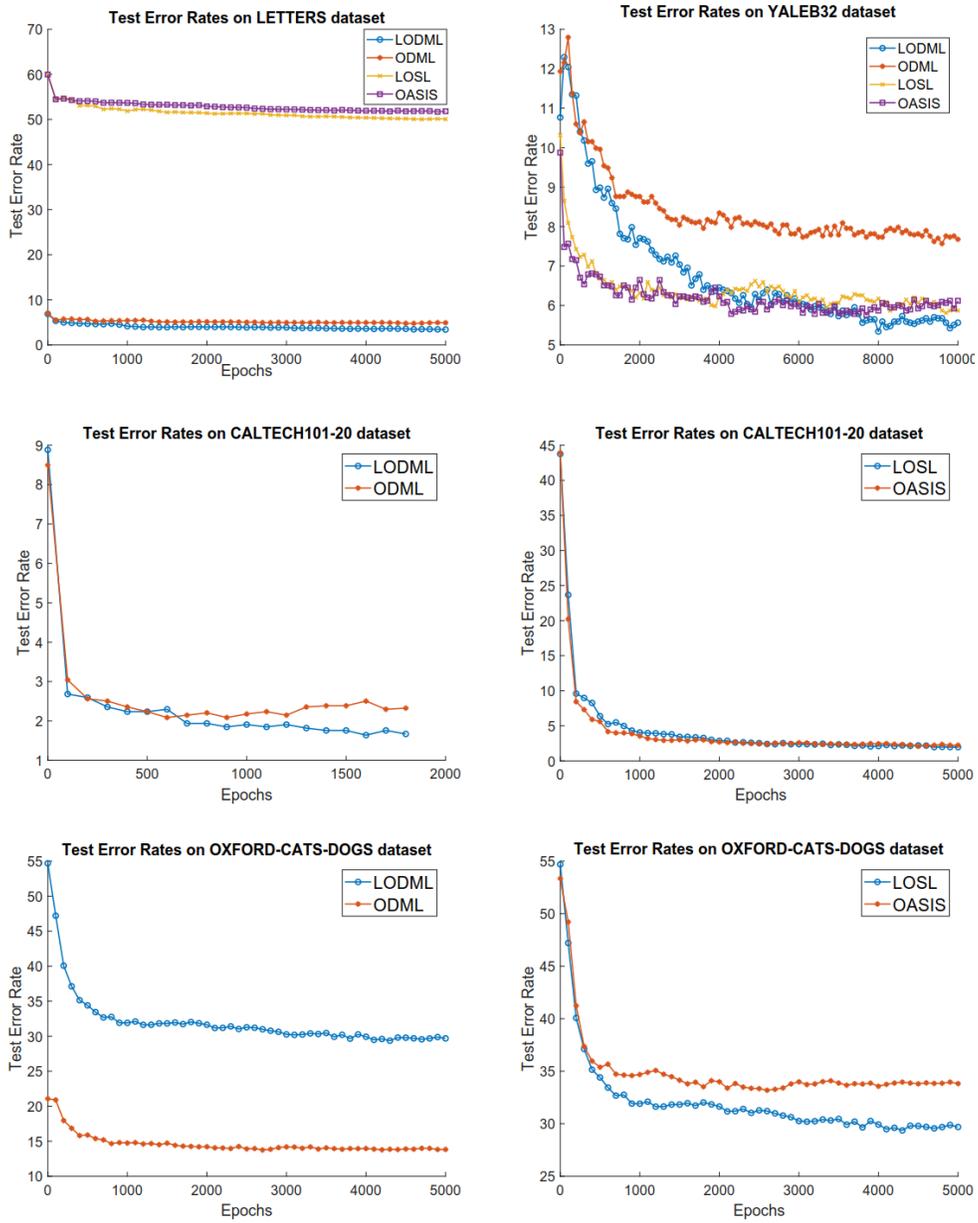

**Figure 4: Comparison of convergence rate of the competing methods on evaluated datasets**

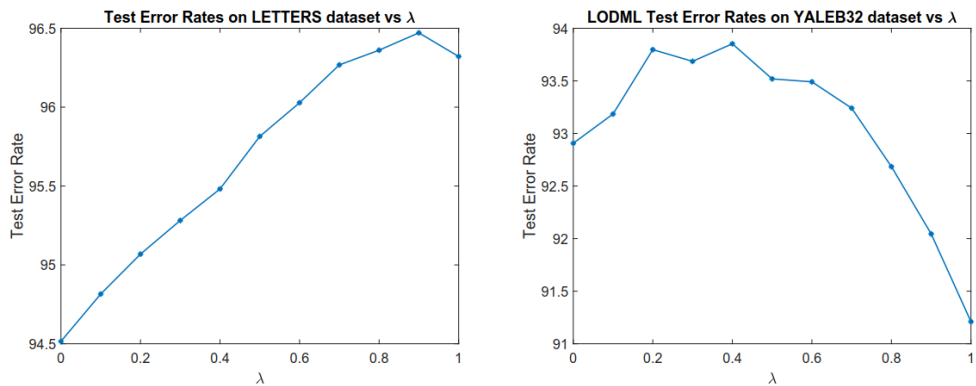

**Figure 5: The influence of the parameter $\lambda$ on the classification rate of the LODML method**
17

In addition, the effect of applying DRP on the LODML method from both the classification rate and execution time aspects is examined. For this purpose, in Figure 6, the predictive performance of LODML with L2ODML is compared on the Caltech101_20, Caltech256_20, CIFAR10 and Oxford_Cats&Dogs datasets. The results confirm that the classification rate the LODML method is almost identical to the L2ODML method and LODML has no significant superiority to L2ODML in terms of classification rate. Also, the running time of these methods are compared together on the Caltech101_20, Caltech256_20, CIFAR10, and Oxford_Cats & Dogs datasets and the results are shown in Figure 7. As the results show, using the DRP, the L2ODML method is scalable with the dimension of data, and its execution time is dramatically better than the LODML method on all of the evaluated datasets. Note that this significant reduction in the computational time is obtained by preserving the discrimination power of the learned metric.

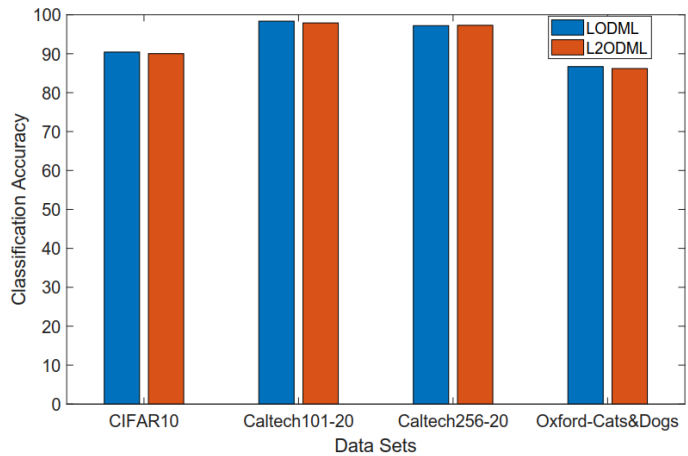

Figure 6: Comparison of the classification rate of LODML with L2ODML

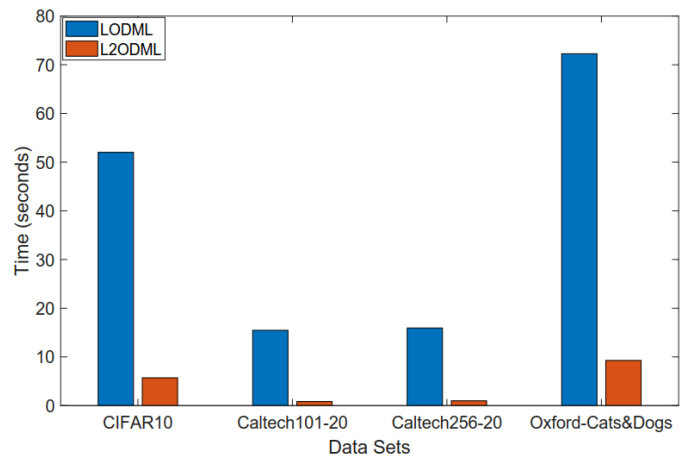

Figure 7: Comparison of the execution time of LODML with L2ODML

## 5. Conclusion and Future Work

This paper presents a scalable local online similarity/distance learning framework based on PA. Each metric in the proposed framework benefits of both the global and local discriminative information. The local component specifically learns the discriminative features of the associated class which increases the discrimination power of the metric. On the other hand, adding the global component to each metric has prevented of overfitting.

The proposed framework can be applied to any global similarity/distance learning algorithm based on PA. This localization significantly improves the predictive performance of the learned measure without increasing the time complexity of the algorithm.

In this research, the proposed framework is applied on both the ODML (Wu et al. 2016) and the OASIS (Chechik et al. 2010) methods and two new local online similarity/distance learning algorithms are produced. These methods are named LOSL and LODML respectively. Experimental results on some challenging datasets from machine vision confirm that the LODML and LOSL methods have a considerably higher performance than the global ones.

To provide scalability with the dimension of data, DRP (Dual Random Projection) is extended for local online learning in the present work. It allows our methods efficiently to be executed on high-dimensional datasets, while maintains their accuracy. In this paper, DRP approach is applied on both the LODML and LOSL methods. The resulting methods are named L2ODML and L2OSL respectively. Experimental results indicate that the execution time of these methods is dramatically better than LODML and LOSL.

Some directions for future work in this area include:

I. Examining the efficiency of the proposed methods in other applications like CBIR (Content Based Information Retrieval).
II. Extension of the proposed methods for semi-supervised learning.
III. Investigating other loss functions in the proposed framework and comparing their efficiency with the Hinge loss function used in this research.

## Appendix A

***Proof* of Theorem 1:**

Note that

$$d_{M_i}(\boldsymbol{p}_k, \boldsymbol{p}_k^+) = (\boldsymbol{p}_k - \boldsymbol{p}_k^+)(\lambda U + \mu L_i)(\boldsymbol{p}_k - \boldsymbol{p}_k^+)^T$$
$$= \lambda \operatorname{tr}(L_i(\boldsymbol{p}_k - \boldsymbol{p}_k^+)(\boldsymbol{p}_k - \boldsymbol{p}_k^+)^T) + \mu \operatorname{tr}(L_i(\boldsymbol{p}_k - \boldsymbol{p}_k^+)(\boldsymbol{p}_k - \boldsymbol{p}_k^+)^T)$$

similarly

$$d_{M_j}(\boldsymbol{p}_k, \boldsymbol{p}_k^-) = (\boldsymbol{p}_k - \boldsymbol{p}_k^-)(\lambda U + \mu L_j)(\boldsymbol{p}_k - \boldsymbol{p}_k^-)^T$$
$$= \lambda \operatorname{tr}(U(\boldsymbol{p}_k - \boldsymbol{p}_k^-)(\boldsymbol{p}_k - \boldsymbol{p}_k^-)^T) + \mu \operatorname{tr}(L_j(\boldsymbol{p}_k - \boldsymbol{p}_k^-)(\boldsymbol{p}_k - \boldsymbol{p}_k^-)^T)$$

Let

$$\boldsymbol{G}^+ = (\boldsymbol{p}_k - \boldsymbol{p}_k^+)(\boldsymbol{p}_k - \boldsymbol{p}_k^+)^T, \quad \boldsymbol{G}^- = (\boldsymbol{p}_k - \boldsymbol{p}_k^-)(\boldsymbol{p}_k - \boldsymbol{p}_k^-)^T$$
$$\boldsymbol{G}_d = \boldsymbol{G}^+ - \boldsymbol{G}^-$$

In this case, the above relations can be written as follows:

$$d_{M_i}(\boldsymbol{p}_k, \boldsymbol{p}_k^+) = \lambda \operatorname{tr}(U\boldsymbol{G}_d) + \mu \operatorname{tr}(L_i \boldsymbol{G}^+)$$
$$d_{M_j}(\boldsymbol{p}_k, \boldsymbol{p}_k^-) = \lambda \operatorname{tr}(U\boldsymbol{G}_d) + \mu \operatorname{tr}(L_j \boldsymbol{G}^-)$$

as a result

$$1 + d_{M_i}(\boldsymbol{p}_k, \boldsymbol{p}_k^+) - d_{M_j}(\boldsymbol{p}_k, \boldsymbol{p}_k^-) = 1 + \lambda \operatorname{tr}(\boldsymbol{U}\boldsymbol{G}^+) + \mu \operatorname{tr}(\boldsymbol{L}_i\boldsymbol{G}^+) - \lambda \operatorname{tr}(\boldsymbol{U}\boldsymbol{G}^-) - \mu \operatorname{tr}(\boldsymbol{L}_j\boldsymbol{G}^-)$$
$$= 1 + \lambda \operatorname{tr}(\boldsymbol{U}\boldsymbol{G}_d) + \mu \operatorname{tr}(\boldsymbol{L}_i\boldsymbol{G}^+) - \mu \operatorname{tr}(\boldsymbol{L}_j\boldsymbol{G}^-)$$

The Lagrangian of the optimization problem (9) is

$$
\begin{aligned}
l = &\tfrac{1}{2}\|\boldsymbol{U} - \boldsymbol{U}^{(k)}\|_F^2 + \tfrac{1}{2}\|\boldsymbol{L}_i - \boldsymbol{L}_i^{(k)}\|_F^2 + \tfrac{1}{2}\|\boldsymbol{L}_j - \boldsymbol{L}_j^{(k)}\|_F^2 + C\xi \\
&+ \tau\big(1 + \lambda \operatorname{tr}(\boldsymbol{U}\boldsymbol{G}_d) + \mu \operatorname{tr}(\boldsymbol{L}_i\boldsymbol{G}^+) - \mu \operatorname{tr}(\boldsymbol{L}_j\boldsymbol{G}^-)\big) - \eta\xi \\
&\eta, \tau \geq 0
\end{aligned}
\quad (A.1)
$$

Setting the partial derivatives of $l$ with respect to $\boldsymbol{U}, \boldsymbol{L}_i, \boldsymbol{L}_j$, and $\xi$ to zero gives

$$\frac{\partial l}{\partial \boldsymbol{U}} = 0 \Rightarrow (\boldsymbol{U} - \boldsymbol{U}^{(k)}) + \tau\lambda \boldsymbol{G}_d = \boldsymbol{0}$$

$$\Rightarrow \boldsymbol{U} = \boldsymbol{U}^{(k)} - \tau\lambda \boldsymbol{G}_d \quad (A.2)$$

$$\frac{\partial l}{\partial \boldsymbol{L}_i} = 0 \Rightarrow \boldsymbol{L}_i - \boldsymbol{L}_i^{(k)} + \tau\mu \boldsymbol{G}^+ = \boldsymbol{0}$$

$$\Rightarrow \boldsymbol{L}_i = \boldsymbol{L}_i^{(k)} - \tau\mu \boldsymbol{G}^+ \quad (A.3)$$

$$\frac{\partial l}{\partial \boldsymbol{L}_j} = 0 \Rightarrow \boldsymbol{L}_j - \boldsymbol{L}_j^{(k)} - \tau\mu \boldsymbol{G}^- = \boldsymbol{0}$$

$$\Rightarrow \boldsymbol{L}_j = \boldsymbol{L}_j^{(k)} + \tau\mu \boldsymbol{G}^- \quad (A.4)$$

$$\frac{\partial l}{\partial \xi} = 0 \Rightarrow C - \tau - \eta = 0 \Rightarrow C = \tau + \eta$$

Since $\eta, \tau \geq 0$, we have

$$0 \leq \tau \leq C \quad (A.5)$$

By plugging (A.2), (A.3), and (A.4) into the Lagrangian problem (A.1), the dual function is obtained as:

$$
\begin{aligned}
D(\tau) = &\tfrac{1}{2}\|\tau\lambda \boldsymbol{G}_d\|_F^2 + \tfrac{1}{2}\|\tau\mu \boldsymbol{G}^+\|_F^2 + \tfrac{1}{2}\|\tau\mu \boldsymbol{G}^-\|_F^2 \\
&+ C\xi + \tau + \lambda\tau \operatorname{tr}\big((\boldsymbol{U}^{(k)} - \tau\lambda \boldsymbol{G}_d)\boldsymbol{G}_d\big) + \tau\mu \operatorname{tr}\big((\boldsymbol{L}_i^{(k)} - \tau\mu \boldsymbol{G}^+)\boldsymbol{G}^+\big) \\
&- \tau\mu \operatorname{tr}\big((\boldsymbol{L}_j^{(k)} + \tau\mu \boldsymbol{G}^-)\boldsymbol{G}^-\big) - \tau\xi - \eta\xi
\end{aligned}
$$

After some simplifications the dual problem will be

$$\underset{\tau}{\text{maximize}}\ D(\tau) = -\frac{\tau^2}{2}(\|\lambda \boldsymbol{G}_d\|_F^2 + \|\mu \boldsymbol{G}^+\|_F^2 + \|\mu \boldsymbol{G}^-\|_F^2)$$
$$+\tau\left(1 + \lambda\,\text{tr}(\boldsymbol{U}^{(k)}\boldsymbol{G}_d) + \mu\,\text{tr}\left(\boldsymbol{L}_i^{(k)}\boldsymbol{G}^+\right) - \mu\,\text{tr}\left(\boldsymbol{L}_j^{(k)}\boldsymbol{G}^-\right)\right) \quad (A.6)$$

$$\text{subject to}\quad 0 \leq \tau \leq C$$

Differentiating the Equation (A.6) with respect to $\tau$ and setting it equal to zero yields:

$$\frac{\partial D}{\partial \tau} = 0 \Rightarrow \tau = \frac{1 + \lambda\,\text{tr}(\boldsymbol{U}^{(k)}\boldsymbol{G}_d) + \mu\,\text{tr}\left(\boldsymbol{L}_i^{(k)}\boldsymbol{G}^+\right) - \mu\,\text{tr}\left(\boldsymbol{L}_j^{(k)}\boldsymbol{G}^-\right)}{\|\lambda \boldsymbol{G}_d\|_F^2 + \|\mu \boldsymbol{G}^+\|_F^2 + \|\mu \boldsymbol{G}^-\|_F^2}$$

Since $0 \leq \tau \leq C$, the optimal value of $\tau$ will be:

$$\tau = \min\left(C, \frac{l^{(k)}(\boldsymbol{p}_k, \boldsymbol{p}_k^+, \boldsymbol{p}_k^-)}{\|\lambda \boldsymbol{G}_d\|_F^2 + \|\mu \boldsymbol{G}^+\|_F^2 + \|\mu \boldsymbol{G}^-\|_F^2}\right) \quad (A.7)$$

where $l^{(k)}(\boldsymbol{p}_k, \boldsymbol{p}_k^+, \boldsymbol{p}_k^-) = \max\left(0, 1 + \lambda\,\text{tr}(\boldsymbol{U}^{(k)}\boldsymbol{G}_d) + \mu\,\text{tr}\left(\boldsymbol{L}_i^{(k)}\boldsymbol{G}^+\right) - \mu\,\text{tr}\left(\boldsymbol{L}_j^{(k)}\boldsymbol{G}^-\right)\right)$ shows the value of loss function in the current model (before the update). ∎

## Acknowledgment
We would like to acknowledge the Machine Learning Lab. in Engineering Faculty of FUM for their kind and technical support.